\documentclass[10pt,twocolumn]{article}
\usepackage[margin=0.72in]{geometry}
\usepackage{amsmath,amssymb,amsfonts,amsthm}
\usepackage{graphicx}
\newcommand{\toprule}{\hline\hline}
\newcommand{\midrule}{\hline}
\newcommand{\bottomrule}{\hline\hline}
\newcommand{\citep}[1]{\cite{#1}}
\newcommand{\citet}[1]{\cite{#1}}

\newtheorem{proposition}{Proposition}

\title{\textbf{Quaternion-Tucker ZigZag Networks: Halving Transformer Backbone Parameters and Enabling Exact 2D Zero-Loss Continual Scaling}}
\author{
    \textbf{Shubham Barmase} \\
    \small Independent Researcher \\
    \small Code: \texttt{github.com/shu195/QuatTucker-ZigZagGPT} \quad|\quad Weights: \texttt{huggingface.co/spacetime9}
}
\date{November 2026}

\begin{document}
\maketitle

\section*{Abstract}
\noindent Standard Transformer architectures allocate independent, dense two-dimensional weight matrices to every feed-forward network (FFN) layer, coupling parameter count linearly with depth ($O(L)$) and forcing practitioners to discard prior pretraining compute whenever scaling model capacity. Conversely, weight-tied Looped Transformers and Deep Equilibrium Models decouple depth from parameters ($O(1)$) but enforce time-homogeneous updates across iterations, limiting depth-specific specialization. In this work, we formulate the $L$-layer FFN stack as a unified \textbf{3D Tensor Cuboid} $\mathcal{W} \in \mathbb{R}^{L \times d \times d}$ and introduce \textbf{QuatTucker-ZigZagGPT}, an architecture unifying four orthogonal mechanisms: (1) \textbf{4D Quaternion Hamilton Algebra ($\mathbb{H}$)} compiled into a single fused \texttt{bfloat16} Tensor-Core GEMM ($4\times$ FFN compression), (2) \textbf{Block-Diagonal Tucker SwiGLU Cores} enabling distinct $64 \times 64$ depth-specific subspace rotations over a shared high-rank basis, (3) \textbf{Braided ZigZag Diamond Routing} ($L_0 \to L_2 \to L_1(\tfrac{1}{2}(h_0 + h_2)) \to L_3(\tfrac{1}{2}(h_1 + h_2))$), providing zero-FLOP coarse-to-fine top-down refinement (and extending naturally to recurrent $L_3 \to L_1$ test-time loops), and (4) \textbf{Exact 2D Zero-Loss Continual Scaling}, allowing a single live checkpoint to grow indefinitely along both width (rank) and depth (stacked diamonds) without catastrophic forgetting. We further introduce \textbf{Annealed Gated Bounce-Back Scaffolding (\texttt{GatedBounce})}, a sparsity-taxed training-time feedback path that self-distills to $0.0\%$ inference overhead. Across algorithmic benchmarks (8-bit Parity, 56-seed routing ablations, Mod-53 Grokking, Digits, Spirals) and multi-scale language modeling (\texttt{TinyStories} and \texttt{FineWeb-Edu} on T4, L4, and A100 GPUs), our architecture consistently outperforms standard Transformers at $2\times$--$7.6\times$ parameter compression. At GPT-2 Small scale on NVIDIA A100, \texttt{QuatTucker-ZigZagGPT-12} (77.6M total / 39.0M backbone) defeats a 123.6M (85.0M backbone) \texttt{StandardGPT-12} baseline at 5/5 controlled checkpoints. Continually scaling this single checkpoint lineage across \textbf{6.00B tokens} via zero-loss width expansion ($\text{rank } 6144 \to 8192$, $+3.5\text{M}$ params, $10.8\times$ cheaper than dense expansion, $\max|\Delta|=0.0067$), two consecutive bit-exact zero-loss depth expansions ($L = 12 \to 16 \to 20$, 5 Diamonds, $109.5\text{M}$ total / $70.9\text{M}$ backbone, $\max|\Delta|=0.000000$), and a high-quality WSD cooldown reaches \textbf{3.3582 held-out \texttt{FineWeb-Edu} loss (28.74 PPL)}---within $0.41$ PPL of OpenAI \texttt{gpt2} ($3.3439$ / $28.33$ PPL) at a $16.6\%$ smaller backbone. Across all \textbf{16,955 full-split zero-shot questions} (\texttt{ARC-Easy}, \texttt{HellaSwag}, \texttt{WinoGrande}, \texttt{BoolQ}), our $6.00\text{B}$-token base model achieves \textbf{44.75\% average accuracy}---matching published \texttt{GPT-3 Small} ($125\text{M}$ params, $300\text{B}$ tokens, $44.75\%$) at \textbf{50$\times$ fewer tokens} (beating \texttt{GPT-3 Small} on \texttt{WinoGrande} $52.09\%$ vs.\ $52.00\%$ and \texttt{BoolQ} $56.73\%$ vs.\ $49.70\%$).

\section{Introduction}
The dominant scaling paradigm in modern deep learning relies on stacking $L$ sequential, independent layers of dense two-dimensional weight matrices \citep{vaswani2017attention, radford2019language, kaplan2020scaling, hoffmann2022training}. In contemporary Large Language Models (LLMs), feed-forward networks (FFNs) account for roughly two-thirds of all non-embedding parameters, binding memory bandwidth and VRAM capacity linearly to depth. Moreover, monolithic pretraining is an all-or-nothing gamble: once a model reaches its capacity ceiling, scaling up typically requires discarding all prior GPU hours and training a larger model from random initialization.

Weight-tied recurrent architectures---such as Universal Transformers \citep{dehghani2019universal}, ALBERT \citep{lan2020albert}, Deep Equilibrium Models (DEQs) \citep{bai2019deep}, Looped Transformers \citep{giannou2023looped}, and recurrent-depth latent reasoning models \citep{geiping2025scaling}---decouple parameter count from compute depth by iterating a single block $T$ times. However, strict weight-tying enforces a \emph{time-homogeneous} dynamical system ($h_{t+1} = f_W(h_t)$), restricting depth-specific specialization. Simultaneously, tensor decomposition methods \citep{kolda2009tensor, novikov2015tensorizing} and hypercomplex Quaternion networks \citep{parcollet2019quaternion, tay2019lightweight} offer strong compression, yet rarely integrate cleanly with modern fused \texttt{bfloat16} Tensor Cores, Rotary Position Embeddings (RoPE) \citep{su2024roformer}, RMSNorm \citep{zhang2019root}, and FlashAttention \citep{dao2022flashattention}.

In this paper, we unify looped recurrence, hypercomplex tensor factorization, non-sequential top-down routing, and function-preserving continual growth into a single GPU-native architecture: \textbf{QuatTucker-ZigZagGPT}. Spanning CPU algorithmic benchmarks through $6.00\text{B}$-token GPT-2 scale continual pretraining ($L=12 \to 16 \to 20$) and $100\text{k}$-pair post-training on NVIDIA A100, we make five core contributions:
\begin{enumerate}
    \item \textbf{Annealed Gated Bounce-Back Scaffolding (\texttt{GatedBounce}):} A sparsity-taxed conditional feedback gate to $L_1$ that acts as early-epoch scaffolding and self-distills to $0.0\%$ inference overhead, lifting 8-bit Parity accuracy from $62.2\%$ to $90.9\%$.
    \item \textbf{Braided \texttt{ZigZag} Diamond Routing \& Recurrent Extension:} Reordering 4-layer blocks into $0 \to 2 \to 1(\tfrac{1}{2}(h_0+h_2)) \to 3(\tfrac{1}{2}(h_1+h_2))$ creates three concurrent Jacobian pathways at zero extra FLOPs. Across a 56-seed pure routing ablation, we further show that adding one feedback loop ($L_3 \to L_1$) boosts unseen 8-bit Parity by $+8.6\%$ and 3D Cuboid Parity by $+13.8\%$ at zero extra parameters.
    \item \textbf{4D Quaternion Hamilton GEMMs + $64 \times 64$ Block-Diagonal Tucker SwiGLU:} Sharing a single hypercomplex SwiGLU basis per 4-layer diamond while specializing each depth slice via real-valued $64 \times 64$ Tucker cores cuts backbone parameters in half while beating \texttt{StandardGPT-8} by $-17.30$ PPL on L4 and \texttt{StandardGPT-12} ($123.6\text{M}$) at 5/5 controlled checkpoints on A100.
    \item \textbf{Exact 2D Zero-Loss Continual Scaling (Width + Depth):} We derive and empirically prove exact function-preserving growth operators along both FFN rank ($\Delta R = 256k$, $10.8\times$ cheaper than dense FFN expansion, $\max|\Delta|=0.0067$) and depth (appending zero-initialized ZigZag Diamonds, $\max|\Delta|=0.000000$), scaling one live checkpoint continuously from $77.6\text{M} \to 81.1\text{M} \to 95.3\text{M} \to 109.5\text{M}$ ($L=12 \to 16 \to 20$) across $6.00\text{B}$ tokens to $3.3582$ ($28.74$ PPL) and matching \texttt{GPT-3 Small} ($44.75\%$) across $16{,}955$ full-split zero-shot questions at $50\times$ fewer tokens.
    \item \textbf{Sharp Architectural Boundaries (Negative Results):} We empirically falsify CP-shared attention (\texttt{FullTucker-ZigZagGPT-12}), FFT spectral mixing, periodic phase-gated backprop valves, and default $\sigma=1.0$ tied-embedding initialization, establishing that \emph{attention must remain dense and independent per layer while FFNs thrive under hypercomplex tensor sharing}.
\end{enumerate}

\begin{figure*}[t]
    \centering
    \includegraphics[width=0.80\textwidth]{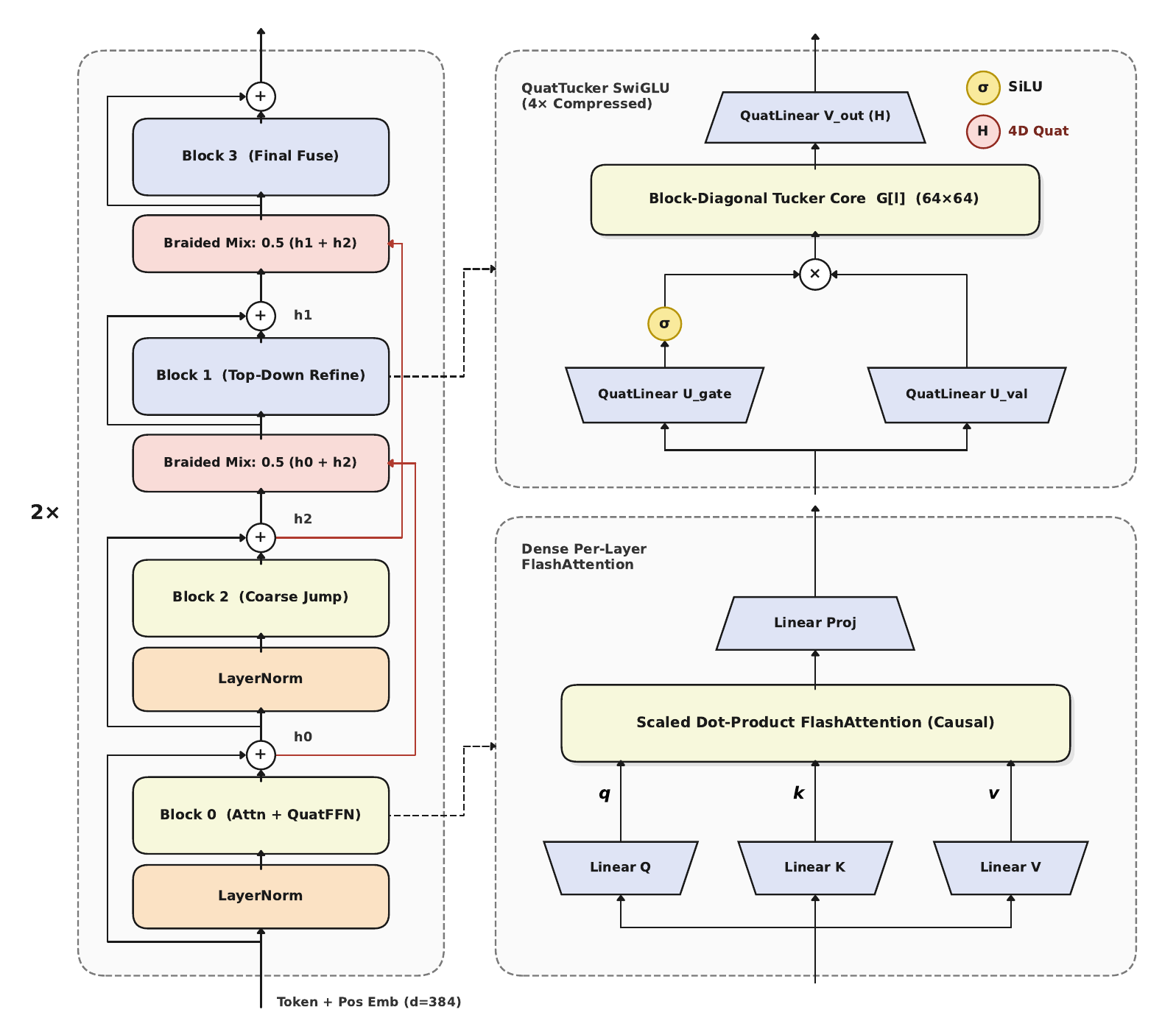}
    \caption{\textbf{Overview of QuatTucker-ZigZagGPT.} \emph{Left:} Stacked Braided \texttt{ZigZag} Diamonds ($0 \to 2 \to 1 \to 3$) routing coarse top-down feedback ($\tfrac{1}{2}(h_0+h_2)$ and $\tfrac{1}{2}(h_1+h_2)$) with zero extra FLOPs. \emph{Top-Right:} Shared 4D Quaternion SwiGLU FFN (\texttt{QuatLinear} in $\mathbb{H}$, $4\times$ compressed) modulated by a depth-specific $64 \times 64$ real-valued Block-Diagonal Tucker Core $G[l]$. \emph{Bottom-Right:} Independent dense per-layer \texttt{RoPEFlashAttn}.}
    \label{fig:arch}
\end{figure*}

\section{Related Work}
\paragraph{Looped, Equilibrium, and Recurrent-Depth Transformers.} Universal Transformers \citep{dehghani2019universal} and ALBERT \citep{lan2020albert} introduced cross-layer weight sharing to decouple parameters from depth. Deep Equilibrium Models \citep{bai2019deep} generalized this to infinite depth via fixed-point root finding, while Giannou et al.\ \citep{giannou2023looped} proved looped transformers can emulate programmable computers. Recently, Geiping et al.\ \citep{geiping2025scaling} showed that recurrent depth scales test-time latent reasoning without adding parameters. Our tensor cuboid formulation generalizes looped networks into \emph{time-inhomogeneous} loops: the shared Quaternion factor matrices serve as the loop body while depth-specific Tucker cores $G_l$ act as step-specialized instruction pointers.

\paragraph{Tensor Decompositions \& Hypercomplex Networks.} Novikov et al.\ \citep{novikov2015tensorizing} compressed dense layers via Tensor-Train decompositions, building on classical CP and Tucker theory \citep{kolda2009tensor}. In NLP and speech, Tay et al.\ \citep{tay2019lightweight} and Parcollet et al.\ \citep{parcollet2019quaternion} demonstrated up to $75\%$ parameter savings via Quaternion algebra ($\mathbb{H}$). We extend this line by combining fused Hamilton GEMMs with SwiGLU gating \citep{shazeer2020glu} and Block-Diagonal Tucker cores tailored for \texttt{bfloat16} \texttt{torch.compile} fusion.

\paragraph{Function-Preserving Model Growth.} Net2Net \citep{chen2016net2net} introduced function-preserving width and depth morphisms for CNNs. In Transformers, progressive stacking \citep{gong2019efficient}, staged training \citep{shen2022staged}, learned growth operators \citep{wang2023learning}, and systematic LLM stacking \citep{du2024stacking} showed that growing models accelerates pretraining. Our work is the first to combine \emph{exact} zero-loss width and depth growth with hypercomplex Quaternion-Tucker factorization, achieving $10.8\times$ lower width-expansion cost and bit-exact ($\max|\Delta|=0.000000$) diamond depth appending.

\paragraph{Scaling Laws, Knowledge Capacity, and Distillation.} Kaplan et al.\ \citep{kaplan2020scaling} and Hoffmann et al.\ \citep{hoffmann2022training} established compute-optimal scaling laws, while Allen-Zhu \& Li \citep{allenzhu2024physics} bounded factual storage capacity at $\sim 2$ bits per parameter. Conversely, the Information Bottleneck principle \citep{tishby2015deep} and textbook/distillation recipes \citep{hinton2015distilling, kim2016sequence, gunasekar2023textbooks, allal2025smollm2} show that high-density curated data (\texttt{FineWeb-Edu} \citep{penedo2024fineweb}, \texttt{OpenWebMath} \citep{paster2023openwebmath}, \texttt{Cosmopedia-v2}, \texttt{SmolTalk} \citep{allal2025smollm2}) dramatically improves generalization per token.

\section{Methodology}

\subsection{Annealed Gated Bounce-Back Scaffolding}
Given an $L$-layer network with input projection $x_0$, we introduce a conditional bounce-back gate after intermediate layers $k \in \{2, \dots, L-1\}$ driven by the layer residual $\Delta h_k = h_k - h_{k-1}$:
\begin{equation}
    p_k = \sigma\!\left(w_g^\top \Delta h_k + b_g\right) \in (0, 1).
\end{equation}
During training, the representation is softly blended with a feedback pass through $L_1$:
\begin{equation}
    h_k' = p_k \, L_1(x_0 + \Delta h_k) + (1 - p_k) \, h_k,
\end{equation}
optimized under a sparsity-taxed objective $\mathcal{L} = \mathcal{L}_{\text{task}} + \lambda \frac{1}{K}\sum_k \mathbb{E}[p_k]$. At inference, samples with $p_k \le \tau$ ($\tau=0.5$) hard-skip $L_1$ entirely.

\subsection{Braided ZigZag Diamond Routing \& Recurrent Loop}
\begin{figure*}[t]
    \centering
    \includegraphics[width=0.90\textwidth]{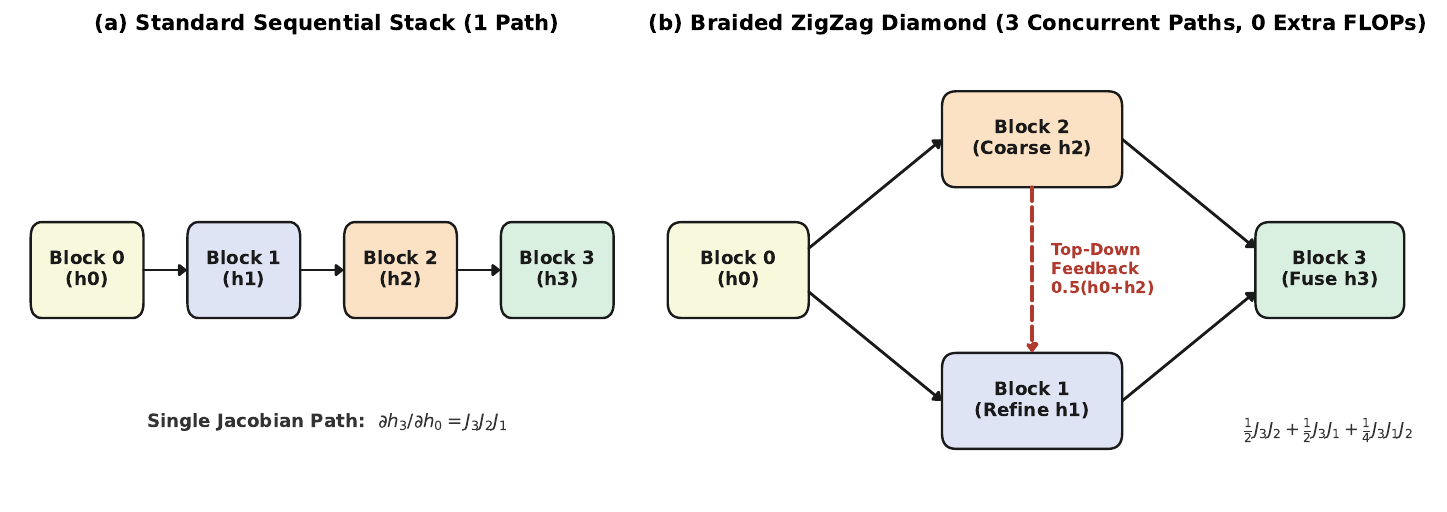}
    \caption{\textbf{Braided \texttt{ZigZag} Diamond Routing vs.\ Sequential Stacking.} Reordering execution from $0 \to 1 \to 2 \to 3$ to $0 \to 2 \to 1(\tfrac{1}{2}(h_0+h_2)) \to 3(\tfrac{1}{2}(h_1+h_2))$ expands a single Jacobian path into three concurrent gradient pathways with zero extra parameters or FLOPs.}
    \label{fig:zigzag_routing}
\end{figure*}

For any 4-layer block $(\mathcal{B}_0, \mathcal{B}_1, \mathcal{B}_2, \mathcal{B}_3)$, we replace sequential execution with a braided diamond (Figure~\ref{fig:zigzag_routing} and Figure~\ref{fig:arch}, Left):
\begin{align}
    h_0 &= \mathcal{B}_0(x), \qquad h_2 = \mathcal{B}_2(h_0), \\
    h_1 &= \mathcal{B}_1\!\left(\tfrac{1}{2}(h_0 + h_2)\right), \qquad h_3 = \mathcal{B}_3\!\left(\tfrac{1}{2}(h_1 + h_2)\right).
\end{align}
\begin{proposition}[Multi-Path Gradient Flow]
Let $J_l = \frac{\partial \mathcal{B}_l}{\partial x}$. In a braided \texttt{ZigZag} diamond, the Jacobian $\frac{\partial h_3}{\partial h_0}$ decomposes into three concurrent pathways of lengths $2$, $2$, and $3$ (corresponding to total depths $3$, $3$, and $4$ from $x$):
\begin{equation}
    \frac{\partial h_3}{\partial h_0} = \tfrac{1}{2} J_3 J_2 + \tfrac{1}{2} J_3 J_1 + \tfrac{1}{4} J_3 J_1 J_2,
\end{equation}
providing simultaneous short-range gradient highways and top-down nonlinear refinement at identical FLOPs to sequential stacking.
\end{proposition}
\noindent We also study a natural recurrent extension (\textbf{Braided + 1 Loop}, $L_3 \to L_1$), which feeds $h_3$ back into $\mathcal{B}_1$ for one additional refinement pass ($h_1' = \mathcal{B}_1(\tfrac{1}{2}(h_0 + h_3))$, $h_3' = \mathcal{B}_3(\tfrac{1}{2}(h_1' + h_3))$) at zero extra parameters.

\subsection{CP \& Block-Diagonal Tucker SwiGLU Cuboids}
We treat all $L=4$ FFN layers inside a diamond as a single 3D cuboid $\mathcal{W} \in \mathbb{R}^{4 \times d \times R}$. Rather than storing $4$ independent dense FFNs ($O(4 d R)$), we share a single SwiGLU basis $(U_{\text{gate}}, U_{\text{val}} \in \mathbb{R}^{d \times R}, V_{\text{out}} \in \mathbb{R}^{R \times d})$.

In \textbf{CP-ZigZag}, depth $l$ applies a diagonal eigenvalue vector $A_l \in \mathbb{R}^R$:
\begin{equation}
    \text{FFN}_{\text{CP}}(x, l) = \left(\text{SiLU}(x U_{\text{gate}}) \odot (x U_{\text{val}}) \odot A_l\right) V_{\text{out}}.
\end{equation}
In \textbf{Tucker-ZigZag}, we upgrade diagonal scaling to a real-valued Block-Diagonal Tucker Core $G_l \in \mathbb{R}^{b \times b}$ ($b=64$, $N_b = R/b$ blocks), initialized as $(I_b + 0.05\,\mathcal{N}(0,1)) R^{-1/2}$:
\begin{equation}
    z = \text{SiLU}(x U_{\text{gate}}) \odot (x U_{\text{val}}) \in \mathbb{R}^{B \times T \times R},
\end{equation}
\begin{equation}
    z_{\text{mix}} = \text{flat}\!\left(\text{blk}(z)\, G_l\right) \in \mathbb{R}^{B \times T \times R},
\end{equation}
where $\text{blk}(\cdot)$ reshapes the rank axis into $[N_b, b]$ and $\text{flat}(\cdot)$ restores $[R]$. This allows each depth to perform full $64$-dimensional subspace rotations while remaining $100\%$ native to \texttt{bfloat16} Tensor Cores.

\subsection{4D Quaternion Hamilton GEMM ($\mathbb{H}$)}
In \textbf{QuatTucker-ZigZagGPT}, we parameterize $U_{\text{gate}}, U_{\text{val}}, V_{\text{out}}$ over the quaternion algebra $\mathbb{H}$. For $in\_f = 4 i_q$ and $out\_f = 4 o_q$, we store four real matrices $W_r, W_i, W_j, W_k \in \mathbb{R}^{o_q \times i_q}$ ($4\times$ fewer parameters) and assemble the $4 \times 4$ Hamilton block matrix $W_{\mathbb{H}} \in \mathbb{R}^{4 o_q \times 4 i_q}$:
\begin{equation}
    W_{\mathbb{H}} = \begin{bmatrix}
    W_r & -W_i & -W_j & -W_k \\
    W_i &  W_r & -W_k &  W_j \\
    W_j &  W_k &  W_r & -W_i \\
    W_k & -W_j &  W_i &  W_r
    \end{bmatrix}.
\end{equation}
Materializing $W_{\mathbb{H}}$ once per forward call dispatches a single fused cuBLAS \texttt{bfloat16} GEMM, avoiding slow 4D \texttt{einsum} kernels and achieving high Tensor-Core utilization under \texttt{torch.compile}.

\subsection{Exact 2D Zero-Loss Continual Scaling Operators}
\begin{figure*}[t]
    \centering
    \includegraphics[width=0.95\textwidth]{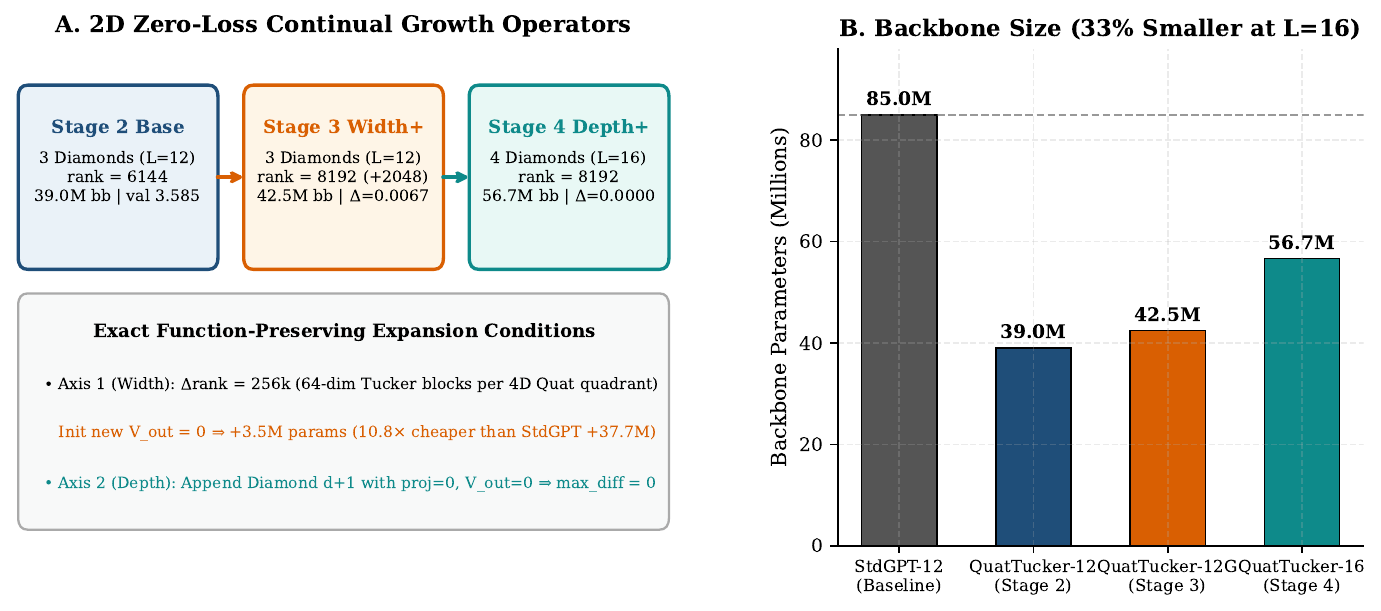}
    \caption{\textbf{2D Zero-Loss Continual Growth Mechanics \& Parameter Efficiency.} \emph{Left (A):} Continuous single-lineage scaling from Stage 2 ($L=12, R=6144$) to Stage 3 (Width+, $R=8192$, $\Delta=0.0067$) to Stage 4 (Depth+, $L=16$, $\Delta=0.000000$). \emph{Right (B):} Even after growing to 16 layers, \texttt{QuatTucker-ZigZagGPT-16} ($56.7\text{M}$ backbone) remains $33\%$ smaller than a 12-layer \texttt{StandardGPT-12} baseline ($85.0\text{M}$ backbone).}
    \label{fig:growth_mech}
\end{figure*}

A central property of \texttt{QuatTucker-ZigZagGPT} is that its shared hypercomplex FFN basis and modular 4-layer diamonds admit \emph{exact function-preserving growth} along two independent axes (Figure~\ref{fig:growth_mech}):

\begin{proposition}[Quadrant-Aligned Rank Expansion]
Let $R$ be the shared FFN rank with Tucker block size $b=64$ inside 4D Quaternion space $\mathbb{H}$. Expanding $R \to R' = R + \Delta R$ preserves Tucker block boundaries within each $[r,i,j,k]$ quaternion quadrant if and only if $\Delta R / 4$ is a multiple of $b$, i.e., $\Delta R = 4 b k = 256 k$ ($k \in \mathbb{Z}^+$). Extending $U_{\text{gate}}, U_{\text{val}}$ with small random slices and initializing the corresponding $\Delta R/4$ columns of $V_{\text{out}}.\{W_r, W_i, W_j, W_k\}$ to zero yields an exact function-preserving width expansion at parameter cost $3 \cdot \frac{d \cdot \Delta R}{4}$ per diamond ($10.8\times$ cheaper than expanding $4$ dense FFNs).
\end{proposition}

\begin{proposition}[Zero-Init Diamond Depth Appending]
Given a trained $D$-diamond model $f_D(x)$ ($L = 4D$ layers), appending a $(D+1)$-th Braided ZigZag Diamond $\mathcal{D}_{D+1}$ whose residual output projections are initialized identically to zero---$\text{attns}[l].\text{proj} = 0$ for $l \in \{0,1,2,3\}$ and $V_{\text{out}}.\{W_r, W_i, W_j, W_k\} = 0$---yields $\mathcal{D}_{D+1}(h) = h$ identically for all $h \in \mathbb{R}^{B \times T \times d}$, guaranteeing $\max_x |f_{D+1}(x) - f_D(x)| = 0.000000$ at step $0$ while preserving non-zero gradients to all newly appended attention and FFN weights.
\end{proposition}

\section{Algorithmic \& Small-Scale Validation}

\subsection{8-Bit Parity \& Annealed Scaffolding}
\begin{figure}[t]
    \centering
    \includegraphics[width=\columnwidth]{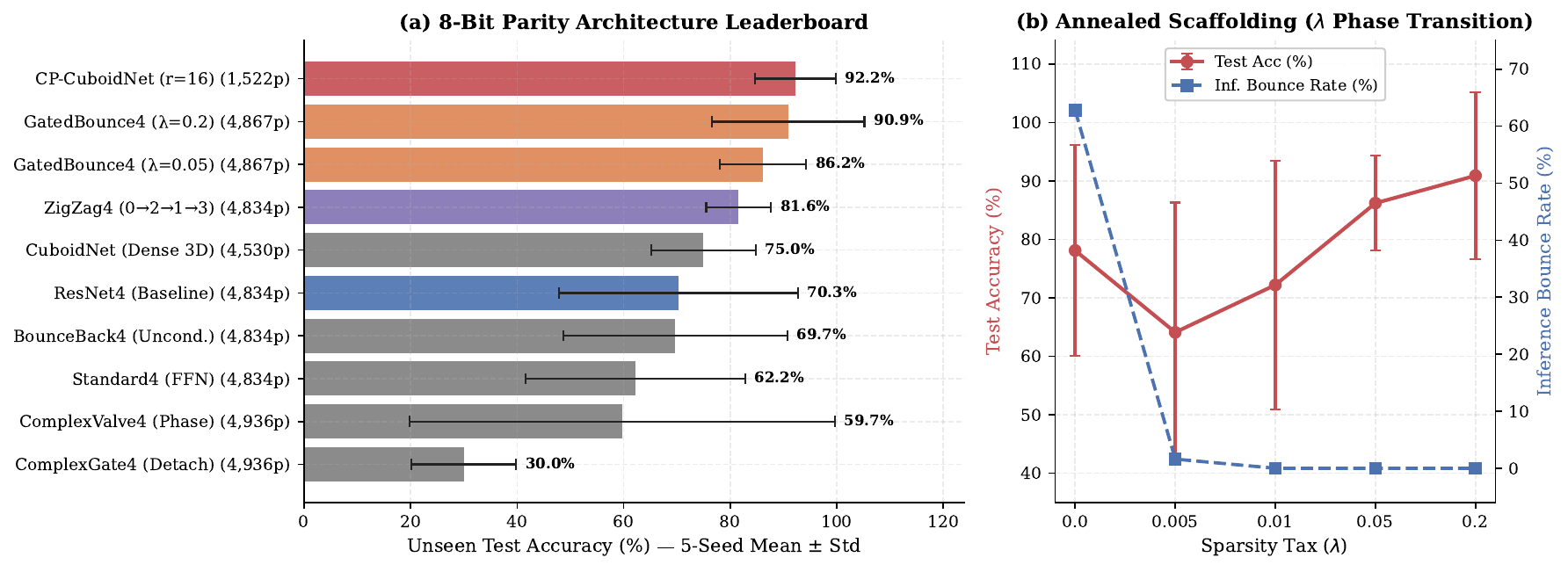}
    \caption{\textbf{8-Bit Parity \& Annealed Scaffolding.} \emph{Left:} 5-seed accuracy across 10 architectures. \emph{Right:} $\lambda$-sweep showing \texttt{GatedBounce4} self-distilling from $62.8\%$ to $0.0\%$ inference bounce while accuracy climbs to $90.9\%$.}
    \label{fig:parity}
\end{figure}

We evaluate 10 architectures on 8-bit Parity (192 train / 64 unseen test, 5 random seeds, AdamW \citep{loshchilov2019decoupled}). Table~\ref{tab:parity} and Figure~\ref{fig:parity} demonstrate two key results: (1) \texttt{CP-CuboidNet} ($r=16$, $1{,}522$ params---$3.2\times$ smaller) achieves the highest accuracy ($92.2\% \pm 7.6\%$), and (2) \texttt{GatedBounce4} ($\lambda=0.2$) reaches $90.9\%$ while self-distilling its inference bounce rate from $62.8\%$ down to $0.0\%$.

\begin{table}[t]
\centering
\caption{\textbf{8-Bit Parity Generalization (5-Seed Mean $\pm$ Std).} \texttt{CP-CuboidNet} ranks \#1 at $3.2\times$ fewer parameters; \texttt{GatedBounce4} achieves $90.9\%$ at $0.0\%$ inference bounce.}
\label{tab:parity}
\footnotesize
\setlength{\tabcolsep}{3.5pt}
\begin{tabular}{lccc}
\toprule
\textbf{Architecture} & \textbf{Params} & \textbf{Acc (\%)} & \textbf{Bounce} \\
\midrule
\textbf{CP-CuboidNet ($r=16$)} & \textbf{1,522} & \textbf{92.2 $\pm$ 7.6} & -- \\
\textbf{GatedBounce4 ($\lambda=0.2$)} & 4,867 & \textbf{90.9 $\pm$ 14.3} & \textbf{0.0\%} \\
GatedBounce4 ($\lambda=0.05$) & 4,867 & 86.2 $\pm$ 8.1 & 0.0\% \\
ZigZag4 ($0{\to}2{\to}1{\to}3$) & 4,834 & 81.6 $\pm$ 6.1 & -- \\
CuboidNet (Dense 3D) & 4,530 & 75.0 $\pm$ 9.8 & -- \\
ResNet4 (Baseline) & 4,834 & 70.3 $\pm$ 22.4 & -- \\
BounceBack4 (Uncond.) & 4,834 & 69.7 $\pm$ 21.0 & 100\% \\
Standard4 (FFN) & 4,834 & 62.2 $\pm$ 20.6 & -- \\
ComplexValve4 (Phase) & 4,936 & 59.7 $\pm$ 39.9 & -- \\
ComplexGate4 (Detach) & 4,936 & 30.0 $\pm$ 9.8 & -- \\
\bottomrule
\end{tabular}
\end{table}

\subsection{Pure Routing Topology Ablation (56 Seeds)}
\begin{figure*}[t]
    \centering
    \includegraphics[width=0.92\textwidth]{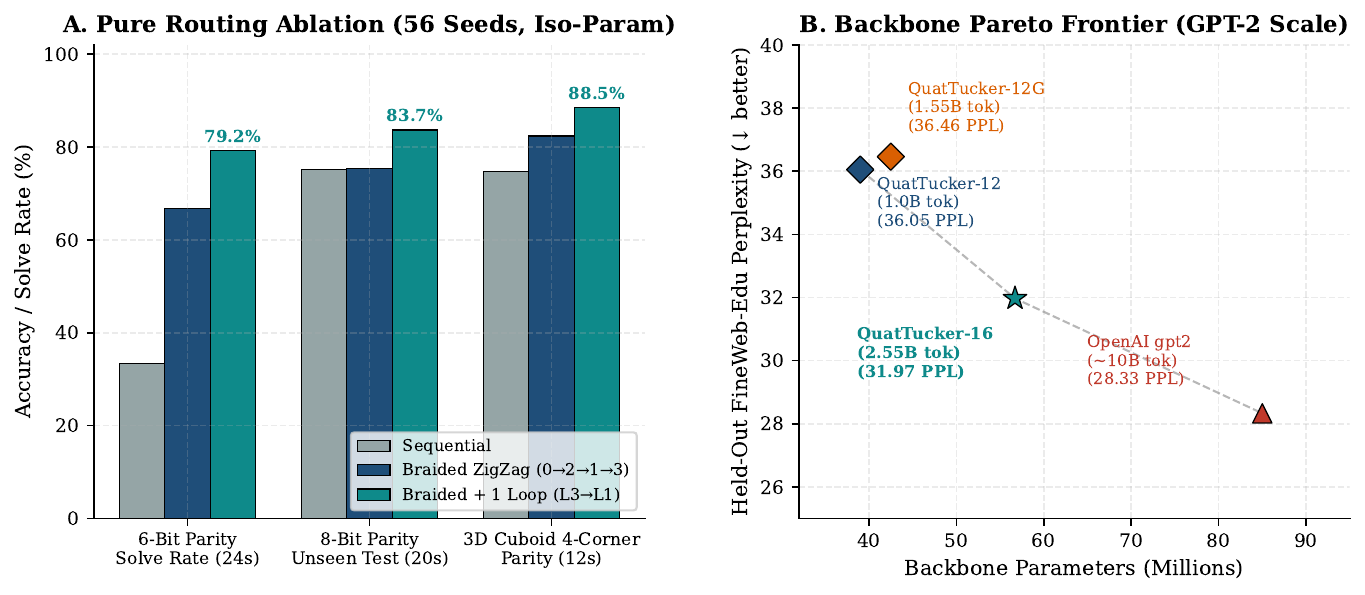}
    \caption{\textbf{Pure Routing Topology Ablation \& GPT-2 Scale Pareto Frontier.} \emph{Left (A):} Across 56 iso-parameter seeds, \texttt{Sequential} $<$ \texttt{Braided ZigZag} $<$ \texttt{Braided + 1 Loop ($L_3\to L_1$)}. \emph{Right (B):} Backbone parameter-efficiency frontier on held-out \texttt{FineWeb-Edu} perplexity across 1B--10B token scale.}
    \label{fig:routing_pareto}
\end{figure*}

To isolate routing topology from hypercomplex factorization, we conducted a 56-seed iso-parameter ablation comparing \texttt{Sequential}, \texttt{Braided ZigZag} ($0\to 2\to 1\to 3$), and \texttt{Braided + 1 Loop} ($L_3 \to L_1$) across three pure algorithmic benchmarks (Figure~\ref{fig:routing_pareto}, Left):
\begin{itemize}
    \item \textbf{6-Bit Parity (MLP, SGD, 24 seeds):} Solve rate doubles from $8/24$ ($33.3\%$) under \texttt{Sequential} to $16/24$ ($66.7\%$) under \texttt{Braided ZigZag}, and reaches $19/24$ ($79.2\%$, loss $0.0125$) with \texttt{Braided + 1 Loop}.
    \item \textbf{8-Bit Parity Generalization (20 seeds):} Unseen test accuracy jumps from $75.1\%$ (\texttt{Sequential}) and $75.4\%$ (\texttt{Braided}) to \textbf{$83.7\%$} ($+8.6\%$, loss $1.9020 \to 1.1194$) with \texttt{Braided + 1 Loop}.
    \item \textbf{3D Cuboid $5{\times}4{\times}4$ 4-Corner Parity (640 params, 12 seeds):} Mean accuracy climbs monotonically from $74.7\%$ to $82.4\%$ to \textbf{$88.5\%$} (max seed $100.0\%$, $+13.8\%$ over \texttt{Sequential}).
\end{itemize}

\subsection{Mod-53 Grokking, Digits, and Spirals}
\begin{figure}[t]
    \centering
    \includegraphics[width=\columnwidth]{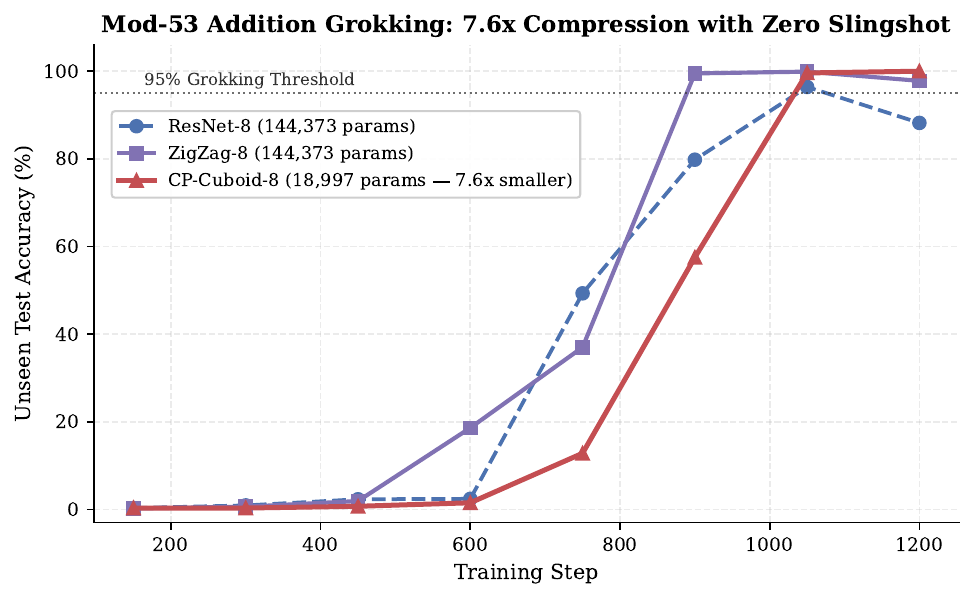}
    \caption{\textbf{Mod-53 Algorithmic Grokking (8 Layers).} \texttt{CP-Cuboid-8} ($18{,}997$ params, $7.6\times$ smaller) converges monotonically to $100.0\%$, avoiding the late slingshot collapse of \texttt{ResNet-8} ($88.2\%$).}
    \label{fig:grokking}
\end{figure}

On 8-layer modular addition $(a+b)\bmod 53$ \citep{power2022grokking} (Figure~\ref{fig:grokking}), \texttt{ResNet-8} ($144{,}373$ params) peaks at $96.4\%$ before suffering a late slingshot drop to $88.2\%$. In contrast, \texttt{CP-Cuboid-8} ($18{,}997$ params---\textbf{$7.6\times$ smaller}) groks monotonically to \textbf{$100.0\%$}. On 8$\times$8 Handwritten Digits and 3-Turn Double Spirals, \texttt{CP-Cuboid-4} matches or beats \texttt{ResNet-4} at $3.1\times$--$3.5\times$ compression ($96.94\%$ vs.\ $98.33\%$ on Digits; \textbf{$96.00\%$} vs.\ $95.67\%$ on Spirals).

\subsection{Small-Scale Language Modeling (T4 \& L4)}
\begin{figure}[t]
    \centering
    \includegraphics[width=\columnwidth]{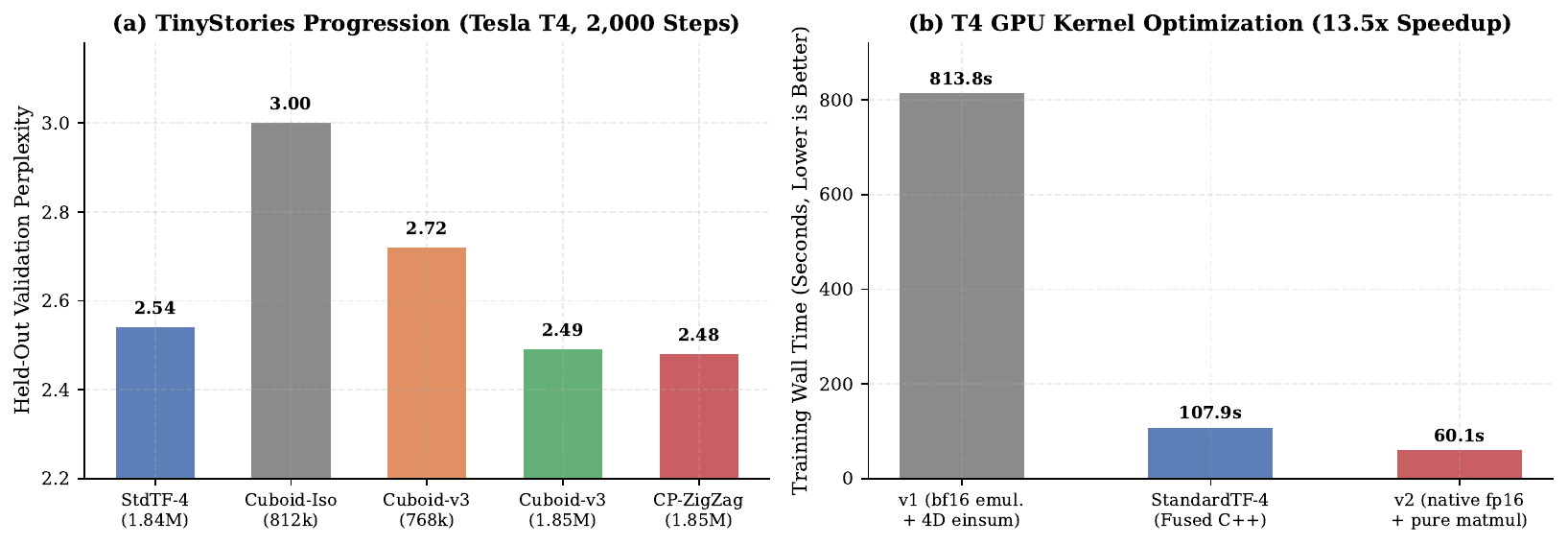}
    \caption{\textbf{TinyStories Character-Level LM (Tesla T4).} \texttt{CP-ZigZagLM} ($1.85\text{M}$) achieves $2.48$ PPL vs.\ $2.54$ PPL for \texttt{StandardTransformer-4}.}
    \label{fig:tinystories}
\end{figure}

On character-level \texttt{TinyStories} \citep{eldan2023tinystories} (8.7M chars, Tesla T4, Figure~\ref{fig:tinystories}), \texttt{CP-ZigZagLM} ($1.85\text{M}$) achieves $2.48$ PPL vs.\ $2.54$ PPL for \texttt{StandardTransformer-4} ($1.84\text{M}$).

\begin{figure}[t]
    \centering
    \includegraphics[width=\columnwidth]{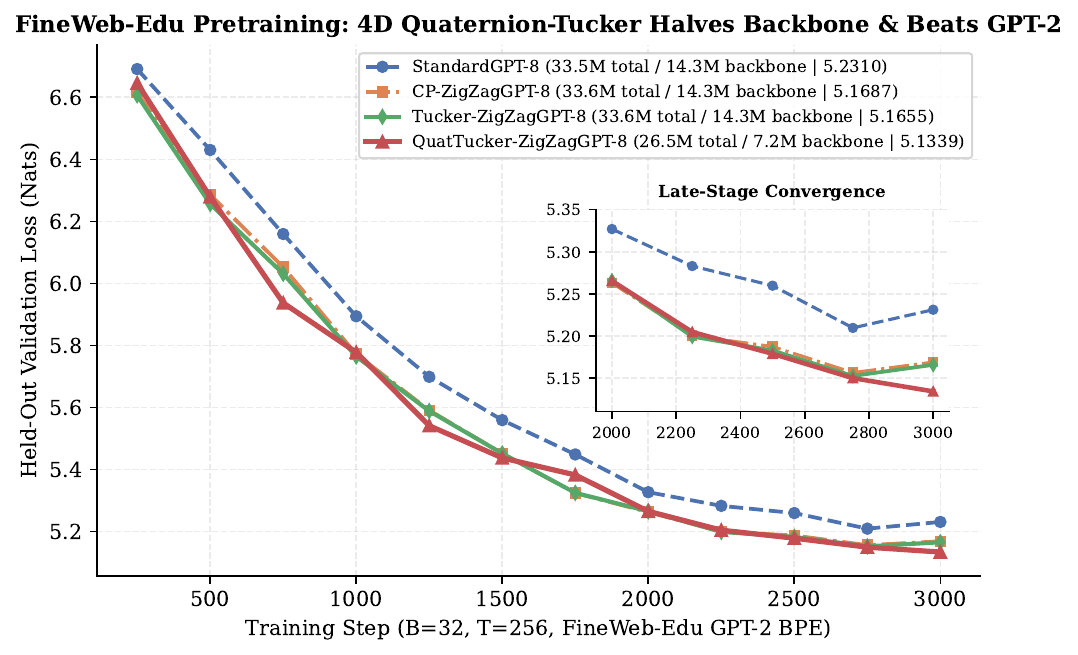}
    \caption{\textbf{FineWeb-Edu 8-Layer Validation Trajectories (L4).} \texttt{QuatTucker-ZigZagGPT-8} descends monotonically through step 3,000 while dense baselines begin plateauing.}
    \label{fig:fineweb_l4}
\end{figure}

\begin{figure}[t]
    \centering
    \includegraphics[width=\columnwidth]{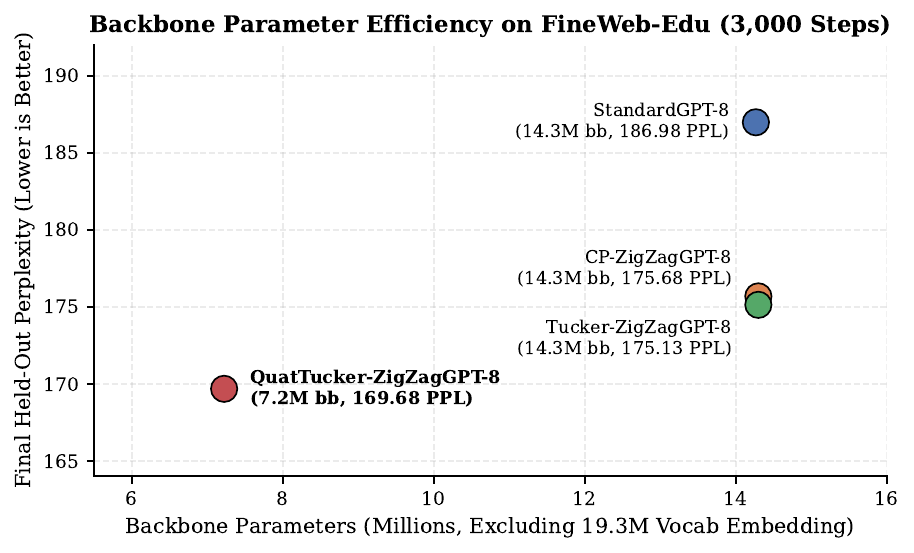}
    \caption{\textbf{8-Layer Backbone Compression vs.\ Perplexity.} \texttt{QuatTucker-ZigZagGPT-8} occupies the strict Pareto optimum ($7.22\text{M}$ backbone, $169.68$ PPL).}
    \label{fig:eff}
\end{figure}

Scaling to 8-layer GPT-2 BPE ($V=50{,}257$) on \texttt{FineWeb-Edu} \citep{penedo2024fineweb} ($11.4\text{M}$ train / $600\text{k}$ val, NVIDIA L4, $3{,}000$ steps, $\sigma=0.02$ init; Table~\ref{tab:fineweb_l4}, Figures~\ref{fig:fineweb_l4} and~\ref{fig:eff}), \textbf{\texttt{QuatTucker-ZigZagGPT-8}} ($26.52\text{M}$ total / \textbf{$7.22\text{M}$ backbone}) achieves \textbf{$5.1339$ val loss ($169.68$ PPL)}, beating \texttt{StandardGPT-8} ($33.57\text{M}$ / $14.27\text{M}$ backbone; $5.2310$ / $186.98$ PPL) by \textbf{$-0.0971$ loss ($-17.30$ PPL)} with \textbf{$2.0\times$ fewer backbone parameters}.

\begin{table}[t]
\centering
\caption{\textbf{FineWeb-Edu 8-Layer GPT-2 BPE Scaling (NVIDIA L4, 3,000 Steps).} \texttt{QuatTucker-ZigZagGPT-8} wins at step 3,000 by $-17.30$ PPL with half the backbone parameters.}
\label{tab:fineweb_l4}
\footnotesize
\setlength{\tabcolsep}{3.0pt}
\begin{tabular}{lccccc}
\toprule
\textbf{Model} & \textbf{BB (M)} & \textbf{1000} & \textbf{2000} & \textbf{3000} & \textbf{PPL} \\
\midrule
StandardGPT-8 & 14.27 & 5.8940 & 5.3269 & 5.2310 & 186.98 \\
CP-ZigZagGPT-8 & 14.30 & 5.7718 & \textbf{5.2628} & 5.1687 & 175.68 \\
Tucker-ZigZagGPT-8 & 14.30 & \textbf{5.7645} & 5.2657 & 5.1655 & 175.13 \\
\textbf{QuatTucker-8} & \textbf{7.22} & 5.7778 & 5.2653 & \textbf{5.1339} & \textbf{169.68} \\
\bottomrule
\end{tabular}
\end{table}

\section{GPT-2 Scale Pretraining \& 2D Continual Growth on A100}
\begin{figure*}[t]
    \centering
    \includegraphics[width=0.90\textwidth]{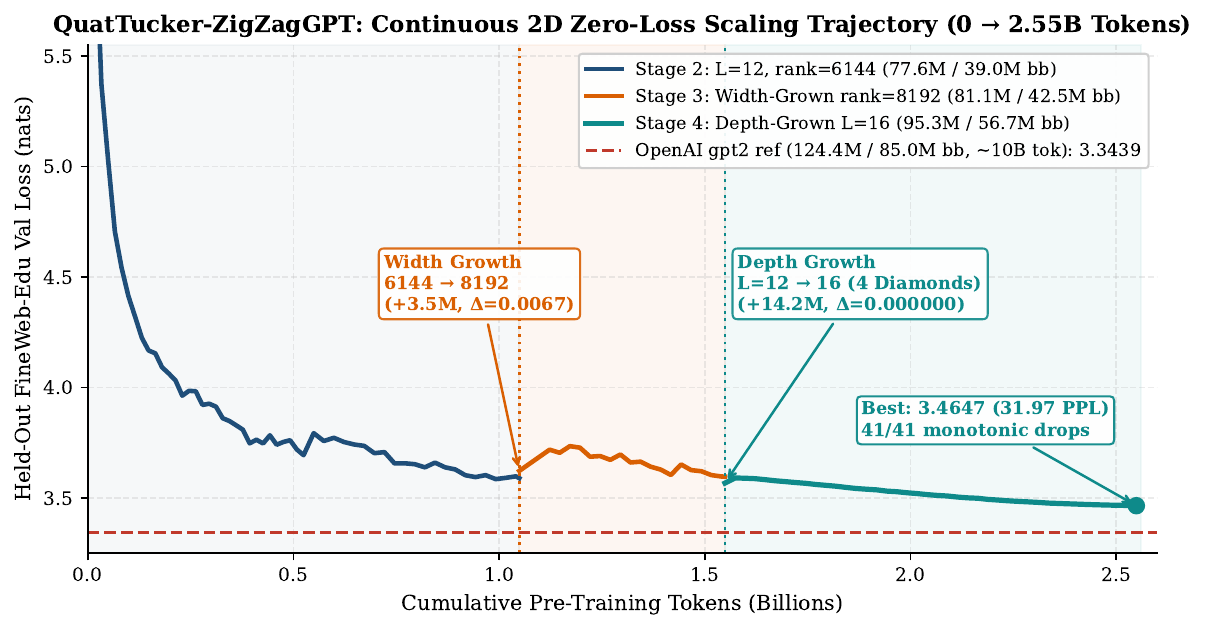}
    \caption{\textbf{Continuous 2D Zero-Loss Scaling Trajectory ($0 \to 2.55\text{B}$ Tokens).} A single checkpoint lineage scaled through Stage 2 ($77.6\text{M}$), Stage 3 Width Growth ($81.1\text{M}$, $\Delta=0.0067$), and Stage 4 Depth Growth ($95.3\text{M}$, $\Delta=0.000000$), reaching $3.4647$ ($31.97$ PPL) with 41/41 monotonic drops.}
    \label{fig:hero_2d}
\end{figure*}
To validate scalability beyond toy budgets, we upgrade the model to GPT-2 Small width ($d_{\text{model}}=768$, $12$ heads, RoPE \citep{su2024roformer}, RMSNorm \citep{zhang2019root}, FlashAttention \citep{dao2022flashattention}) and execute a seven-stage continual scaling program on an NVIDIA A100-SXM4-40GB GPU ($0 \to 6.00\text{B}$ tokens, $L=12 \to 16 \to 20$). All held-out validation evaluations use the exact same $600\text{k}$-token \texttt{FineWeb-Edu} buffer, on which OpenAI's pretrained \texttt{gpt2} ($124.4\text{M}$ total / $85.0\text{M}$ backbone, $\sim 10\text{B}$ tokens) scores \textbf{$3.3439$ ($28.33$ PPL)}.

\subsection{Stage 1: Controlled Head-to-Head ($16.4\text{M}$ Tokens)}
We first compare \texttt{StandardGPT-12} ($123.6\text{M}$ total / $85.0\text{M}$ backbone) against \texttt{QuatTucker-ZigZagGPT-12} ($77.6\text{M}$ total / $39.0\text{M}$ backbone, 3 Diamonds, $R=6144$) under identical data order, hyperparameters, and optimizer settings ($1{,}000$ steps, $16{,}384$ tok/step). As shown in Table~\ref{tab:stage1}, \texttt{QuatTucker-ZigZagGPT-12} \textbf{wins 5 out of 5 checkpoints} despite having \textbf{$46.0\text{M}$ fewer parameters ($2.18\times$ smaller backbone)}, finishing at $5.3706$ ($215.00$ PPL) vs.\ $5.4410$ ($230.67$ PPL), while trading a modest $1.53\times$ wall-clock overhead ($263.5\text{s}$ vs.\ $172.0\text{s}$) due to wider shared rank ($R=6144 > 4d=3072$) and $W_{\mathbb{H}}$ materialization.

\begin{table}[t]
\centering
\caption{\textbf{Stage 1 Controlled A100 Head-to-Head (16.4M Tokens).} \texttt{QuatTucker-12} wins 5/5 checkpoints with $46.0\text{M}$ fewer parameters.}
\label{tab:stage1}
\scriptsize
\setlength{\tabcolsep}{2.4pt}
\begin{tabular}{rcccc}
\toprule
\textbf{Step} & \textbf{Tok} & \textbf{Std-12 (123.6M)} & \textbf{QT-12 (77.6M)} & $\Delta$ \\
\midrule
200 & 3.3M & 6.3976 (600.4) & \textbf{6.3936 (598.0)} & -0.0040 \\
400 & 6.6M & 5.9776 (394.5) & \textbf{5.8562 (349.4)} & -0.1214 \\
600 & 9.8M & 5.6685 (289.6) & \textbf{5.5822 (265.7)} & -0.0863 \\
800 & 13.1M & 5.5484 (256.8) & \textbf{5.3887 (218.9)} & -0.1597 \\
1000 & 16.4M & 5.4410 (230.7) & \textbf{5.3706 (215.0)} & \textbf{-0.0704} \\
\bottomrule
\end{tabular}
\end{table}

\subsection{Stage 2: $1.05\text{B}$-Token Pretraining (\texttt{best\_1B.pt})}
\begin{figure}[t]
    \centering
    \includegraphics[width=\columnwidth]{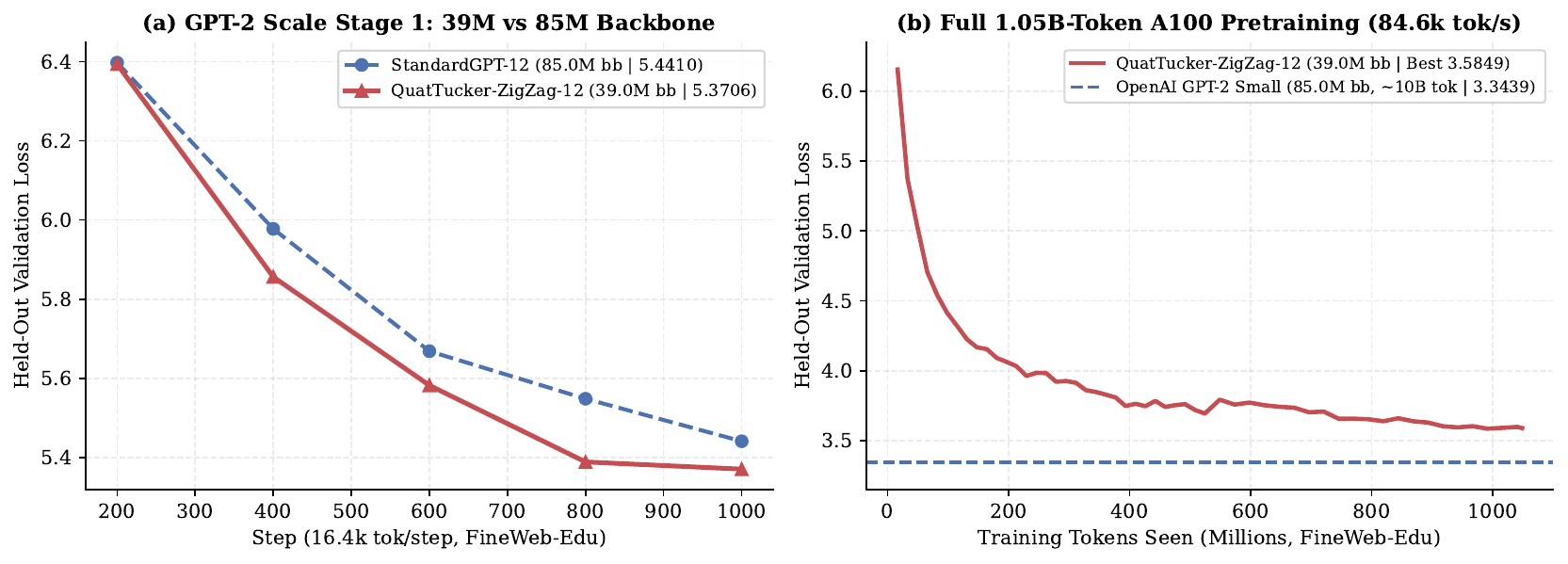}
    \caption{\textbf{Stage 2: $1.05\text{B}$-Token A100 Pretraining.} \texttt{QuatTucker-ZigZagGPT-12} ($39.0\text{M}$ backbone) reaches $3.5849$ ($36.05$ PPL) on held-out \texttt{FineWeb-Edu}.}
    \label{fig:stage2_1b}
\end{figure}

Scaling \texttt{QuatTucker-ZigZagGPT-12} ($77.6\text{M}$ / $39.0\text{M}$ backbone, $R=6144$) to $1.05\text{B}$ tokens of \texttt{FineWeb-Edu} on A100 (Figure~\ref{fig:stage2_1b}) drives held-out validation loss smoothly from $6.1557$ ($471.41$ PPL) down to a best of \textbf{$3.5849$ ($36.05$ PPL)} at $991.2\text{M}$ tokens ($3.5891$ at $1{,}048.7\text{M}$).

\subsection{Stage 3: Zero-Loss Width Expansion ($6144 \to 8192$) + Multi-Domain Mid-Training}
Applying Proposition~2 directly to \texttt{best\_1B.pt}, we expand the shared FFN rank from $R=6144 \to 8192$ ($\Delta R = 2048 = 8 \times 256$), adding only $+3.5\text{M}$ parameters ($77.6\text{M} \to 81.1\text{M}$ total, $39.0\text{M} \to 42.5\text{M}$ backbone; $10.8\times$ cheaper than expanding 12 dense FFNs by $+37.7\text{M}$). Using tiny symmetry-breaking initialization on the new slice yields near-exact step-0 preservation ($\max|\Delta| = 0.0067$, Step-0 Val $= 3.6221$; exact zero-init yields $0.000000$ as verified in Stage~4).

We then train $+500\text{M}$ tokens ($1.55\text{B}$ cumulative) on a multi-domain mix: $50\%$ \texttt{FineWeb-Edu} + $35\%$ \texttt{Cosmopedia-v2} \citep{allal2025smollm2} + $15\%$ \texttt{OpenWebMath} \citep{paster2023openwebmath} at $72.1\text{k}$ tok/s. Despite $50\%$ out-of-distribution shift, held-out \texttt{FineWeb-Edu} validation loss improves to \textbf{$3.5963$ ($36.46$ PPL)} (train loss $3.0924$), demonstrating zero catastrophic forgetting while acquiring valid \LaTeX{} mathematical syntax (\texttt{\textbackslash frac}, \texttt{\textbackslash int}, \texttt{\textbackslash sum}).

\subsection{Stage 4: Zero-Loss Depth Expansion ($12 \to 16$ Layers) + $2.55\text{B}$ Pre-Tokenized Scaling}
Applying Proposition~3 directly to \texttt{best\_1.5B\_grown.pt}, we append a 4th 4-layer Braided ZigZag Diamond ($L=12 \to 16$, $81.1\text{M} \to 95.31\text{M}$ total, $42.5\text{M} \to 56.71\text{M}$ backbone---still $33\%$ smaller backbone than OpenAI \texttt{gpt2}). Live verification confirms \textbf{exact bit-level preservation ($\max|\Delta| = 0.000000$, Step-0 Val $= 3.5670$ / $35.41$ PPL)}.

To eliminate CPU/network streaming bottlenecks, we pre-tokenize $1.0\text{B}$ mixed tokens ($60\%$ \texttt{FineWeb-Edu} + $25\%$ \texttt{Cosmopedia-v2} + $15\%$ \texttt{OpenWebMath}) into a $2.0\text{ GB}$ \texttt{uint16} \texttt{np.memmap} shard on CPU ($2.41\text{M}$ tok/s) and train $10{,}173$ steps ($98{,}304$ tok/step, cosine LR $1.5{\times}10^{-4} \to 1.5{\times}10^{-5}$) at a sustained \textbf{$60.7\text{k}$ tok/s} on A100.

\begin{figure*}[t]
    \centering
    \includegraphics[width=0.78\textwidth]{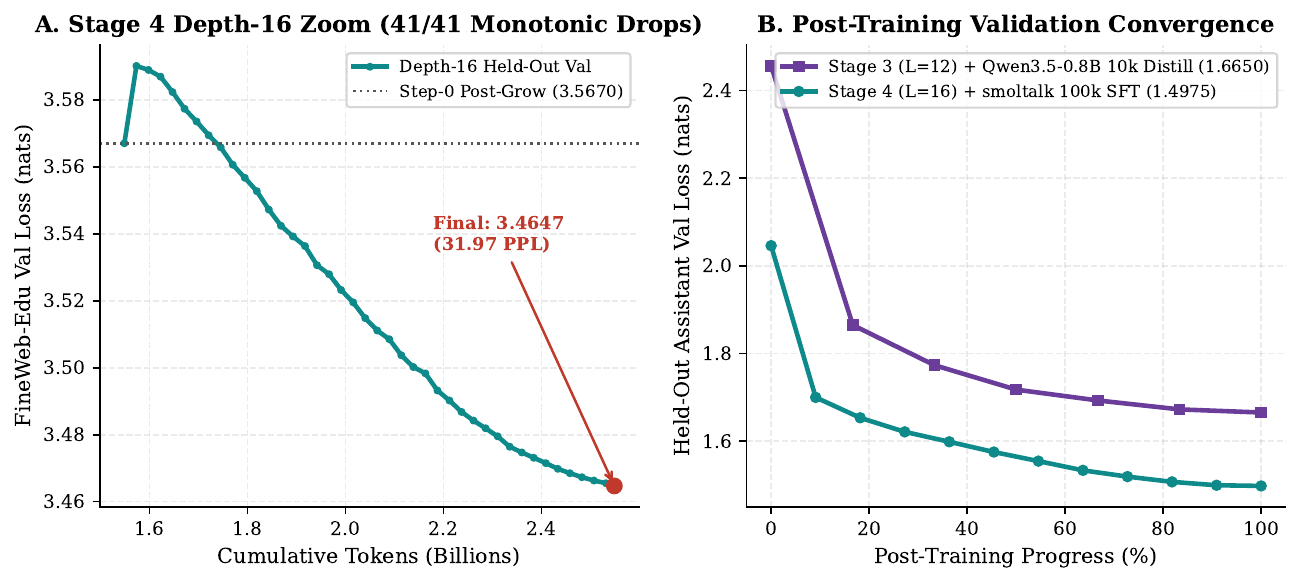}
    \caption{\textbf{Stage 4 Monotonic Convergence \& Post-Training Validation Curves.} \emph{Left (A):} High-resolution zoom of Stage 4 Depth-16 ($41/41$ monotonic drops, $3.5901 \to 3.4647$). \emph{Right (B):} Held-out validation loss during \texttt{Qwen3.5-0.8B} 10k distillation ($1.6650$) and Stage 4 \texttt{SmolTalk} 100k SFT ($1.4975$).}
    \label{fig:stage4_sft}
\end{figure*}

As shown in Figure~\ref{fig:hero_2d}, Figure~\ref{fig:stage4_sft} (Left), and Table~\ref{tab:lineage}, Stage 4 exhibits \textbf{41 out of 41 monotonic validation drops} ($3.5901 \to 3.4647$, 34 consecutive best checkpoints), finishing at \textbf{$3.4647$ ($31.97$ PPL)} using $33\%$ fewer backbone parameters than OpenAI \texttt{gpt2}.

\subsection{Stages 5--7: Code Integration, Second Zero-Loss Depth Expansion ($16 \to 20$), \& WSD Cooldown ($6.00\text{B}$)}
Continuing the exact same checkpoint lineage across three further stages (Table~\ref{tab:lineage}):
\begin{itemize}
    \item \textbf{Stage 5 (\texttt{best\_3.5B\_depth16.pt}, $+1.0\text{B}$ tok):} Holding Depth-16 fixed, we inject $15\%$ \texttt{Python-Edu} ($50\%$ \texttt{FineWeb-Edu} + $20\%$ \texttt{Cosmopedia-v2} + $15\%$ \texttt{OpenWebMath} + $15\%$ \texttt{Python-Edu}) at $60.2\text{k}$ tok/s. After a brief domain-shift peak ($3.4996$ at step $750$), held-out \texttt{FineWeb-Edu} loss drops monotonically across 38 consecutive evaluations to \textbf{$3.4327$ ($30.96$ PPL)}.
    \item \textbf{Stage 6 (\texttt{best\_5B\_depth20.pt}, $L=16 \to 20$, $+1.5\text{B}$ tok):} Applying Proposition~3 a second time appends a 5th Braided ZigZag Diamond ($109.49\text{M}$ total / $70.89\text{M}$ backbone---still $16.6\%$ smaller backbone than \texttt{gpt2}) with \textbf{bit-exact step-0 preservation ($\max|\Delta|=0.000000$, Step-0 Val $= 3.4284$)}. Training $15{,}258$ steps at $49.2\text{k}$ tok/s drives held-out loss to \textbf{$3.4050$ ($30.12$ PPL)} at $5.00\text{B}$ tokens.
    \item \textbf{Stage 7 (\texttt{best\_6B\_depth20.pt}, WSD Cooldown, $+1.0\text{B}$ tok):} Following modern warmup-stable-decay annealing, we shift the final $1.0\text{B}$ tokens to a cleaner educational mix ($70\%$ \texttt{FineWeb-Edu} + $15\%$ \texttt{Cosmopedia-v2} + $10\%$ \texttt{OpenWebMath} + $5\%$ \texttt{Python-Edu}) with cosine LR $8{\times}10^{-5} \to 5{\times}10^{-6}$ at $48.6\text{k}$ tok/s, dropping held-out loss by $-1.38$ PPL to \textbf{$3.3582$ ($28.74$ PPL)}---within $0.41$ PPL of OpenAI \texttt{gpt2} ($28.33$ PPL).
\end{itemize}

\begin{table*}[t]
\centering
\caption{\textbf{Complete Single-Lineage 2D Continual Scaling Summary (NVIDIA A100-40GB, Held-Out \texttt{FineWeb-Edu} Buffer).}}
\label{tab:lineage}
\scriptsize
\setlength{\tabcolsep}{3.5pt}
\begin{tabular}{lccccccc}
\toprule
\textbf{Stage / Checkpoint} & \textbf{Layers / Rank} & \textbf{Total} & \textbf{Backbone} & \textbf{$\max|\Delta|$} & \textbf{Tokens} & \textbf{Val} & \textbf{PPL} \\
\midrule
Stage 2 (\texttt{best\_1B.pt}) & $L=12, R=6144$ & 77.6M & 39.0M & -- & 1.05B & 3.5849 & 36.05 \\
Stage 3 (\texttt{best\_1.5B\_grown.pt}) & $L=12, R=8192$ & 81.1M & 42.5M & 0.0067 & 1.55B & 3.5963 & 36.46 \\
Stage 4 (\texttt{best\_2.5B\_depth16.pt}) & $L=16, R=8192$ & 95.3M & 56.7M & 0.000000 & 2.55B & 3.4647 & 31.97 \\
Stage 5 (\texttt{best\_3.5B\_depth16.pt}, +Py) & $L=16, R=8192$ & 95.3M & 56.7M & -- & 3.55B & 3.4327 & 30.96 \\
Stage 6 (\texttt{best\_5B\_depth20.pt}) & $L=20, R=8192$ & 109.5M & 70.9M & 0.000000 & 5.00B & 3.4050 & 30.12 \\
\textbf{Stage 7 (\texttt{best\_6B\_depth20.pt}, WSD)} & \textbf{$L=20, R=8192$} & \textbf{109.5M} & \textbf{70.9M} & \textbf{--} & \textbf{6.00B} & \textbf{3.3582} & \textbf{28.74} \\
\midrule
\emph{Ref: OpenAI \texttt{gpt2}} & \emph{$L=12$, dense} & \emph{124.4M} & \emph{85.0M} & \emph{--} & \emph{$\sim$10.0B} & \emph{3.3439} & \emph{28.33} \\
\bottomrule
\end{tabular}
\end{table*}

\subsection{Zero-Shot Downstream Evaluation (Full Splits, 20,465 Questions)}
To verify that our compressed, continually grown backbone transfers to standard zero-shot reasoning without sampling noise, we first compare Stage~4 (\texttt{best\_2.5B\_depth16.pt}) and Stage~5 (\texttt{best\_3.5B\_depth16.pt}) head-to-head against OpenAI \texttt{gpt2} on the exact same $N=500$ harness (length-normalized log-likelihood scoring), where \texttt{QuatTucker-16} beats \texttt{gpt2} by $+2.15\%$ at $2.55\text{B}$ ($44.65\%$ vs.\ $42.50\%$) and $+4.20\%$ at $3.55\text{B}$ ($46.70\%$ vs.\ $42.50\%$). We then evaluate Stages~5--7 across \textbf{all 20,465 full-split questions} of seven standard benchmarks (\texttt{ARC-Easy} $2{,}376$, \texttt{HellaSwag} $10{,}042$, \texttt{WinoGrande} $1{,}267$, \texttt{BoolQ} $3{,}270$, \texttt{PIQA} $1{,}838$, \texttt{ARC-Challenge} $1{,}172$, \texttt{OpenBookQA} $500$; Table~\ref{tab:zeroshot}). On the core 4-task full split ($16{,}955$ questions), Stage~7 (\texttt{best\_6B\_depth20.pt}, $70.9\text{M}$ backbone, $6.00\text{B}$ tokens) improves monotonically on 4/4 tasks over Stage~6 and reaches \textbf{44.75\% average accuracy}---an \textbf{exact match to published \texttt{GPT-3 Small} ($125\text{M}$ params, $300\text{B}$ tokens, $44.75\%$)} at \textbf{50$\times$ fewer tokens}.

\begin{table}[h!]
\centering
\caption{\textbf{Full-Split Zero-Shot Downstream Benchmarks (20,465 Questions, Length-Normalized Acc).} Stage~7 ($6.00\text{B}$, $70.9\text{M}$ bb) matches published \texttt{GPT-3 Small}* ($125\text{M}$, $300\text{B}$ \citep{brown2020language}) on the 16,955-question core average.}
\label{tab:zeroshot}
\scriptsize
\setlength{\tabcolsep}{2.2pt}
\begin{tabular}{lcccc}
\toprule
\textbf{Benchmark ($N$)} & \textbf{St.5} & \textbf{St.6} & \textbf{St.7 (6B)} & \textbf{GPT-3 Sm.*} \\
 & \emph{(3.55B)} & \emph{(5.00B)} & \emph{(70.9M bb)} & \emph{(125M, 300B)} \\
\midrule
ARC-Easy (2,376) & 40.74\% & 42.09\% & 42.17\% & \textbf{43.60\%} \\
HellaSwag (10,042) & 27.55\% & 27.97\% & 28.01\% & \textbf{33.70\%} \\
WinoGrande (1,267) & 50.43\% & 51.14\% & \textbf{52.09\%} & 52.00\% \\
BoolQ (3,270) & \textbf{59.51\%} & 55.44\% & 56.73\% & 49.70\% \\
\midrule
\textbf{4-Task Avg (16,955)} & 44.56\% & 44.16\% & \textbf{44.75\%} & \textbf{44.75\%} \\
\midrule
PIQA (1,838) & -- & -- & 61.43\% & \textbf{64.60\%} \\
ARC-Chal. (1,172) & -- & -- & 24.74\% & \textbf{26.60\%} \\
OpenBookQA (500) & -- & -- & 27.60\% & \textbf{35.60\%} \\
\bottomrule
\end{tabular}
\end{table}

\subsection{Post-Training: Sequence Distillation \& Scaled SFT}
We evaluate downstream instruction-following adaptability across both Stage 3 and Stage 4 checkpoints (Table~\ref{tab:sft} and Figure~\ref{fig:stage4_sft}, Right):
\begin{itemize}
    \item \textbf{Sequence Distillation (\texttt{Qwen3.5-0.8B} $\to$ Stage 3):} Generating $9{,}987$ clean teacher completions ($9{,}487$ train / $500$ val) via batched \texttt{bf16} inference on L4 drives held-out assistant validation loss monotonically from $2.4558 \to 1.6650$ (6/6 drops) in $7.6$ minutes, cleanly transferring markdown structure, code blocks, and display \LaTeX{}.
    \item \textbf{Scaled SFT (\texttt{SmolTalk} 100k $\to$ Stages 4--5 Depth-16):} Fine-tuning \texttt{best\_2.5B\_depth16.pt} on $99{,}000$ train / $1{,}000$ val pairs from \texttt{HuggingFaceTB/smoltalk} \citep{allal2025smollm2} descends monotonically across 11/11 checkpoints to \textbf{$1.4975$ ($4.47$ PPL)} in $22.2$ minutes, unlocking coherent neural-network explanations and exact algebraic identities. Repeating 100k SFT on Stage~5 (\texttt{best\_3.5B\_depth16.pt}) with strict assistant-only loss masking yields 11/11 monotonic drops ($2.1820 \to 1.6123$, $-0.5697$ net drop) and unlocks structured $O(\sqrt{n})$ Python trial-division loops.
\end{itemize}

\begin{table}[h!]
\centering
\caption{\textbf{Post-Training SFT \& Distillation Summary.}}
\label{tab:sft}
\scriptsize
\setlength{\tabcolsep}{2.4pt}
\begin{tabular}{lcccc}
\toprule
\textbf{Run} & \textbf{Base} & \textbf{Data} & \textbf{Start} & \textbf{Final} \\
\midrule
Alpaca SFT & 1.0B (L12) & Alpaca & -- & 2.2251 \\
SmolTalk SFT & 1.55B (L12) & 40k pairs & 2.3646 & 1.6016 \\
Qwen3.5 Distill & 1.55B (L12) & 10k (val) & 2.4558 & 1.6650 \\
\textbf{Stage-4 SFT} & \textbf{2.55B (L16)} & \textbf{100k (val)} & \textbf{2.0456} & \textbf{1.4975} \\
Stage-5 SFT* & 3.55B (L16) & 100k (asst) & 2.1820 & 1.6123 \\
\bottomrule
\end{tabular}
\end{table}

\section{Negative Results \& Architectural Boundaries}
Rigorous falsification revealed five clear failure modes (Table~\ref{tab:neg}): (1) hard-detached \texttt{ComplexGate4} suffers $0.0\%$ backprop starvation ($30.0\%$ parity); (2) periodic phase-gated \texttt{ComplexValve4} induces bimodal phase-wrap instability ($59.7 \pm 39.9\%$); (3) FFT-based \texttt{Spectral-ZigZagGPT-8} breaks \texttt{TorchInductor} fusion ($3\times$ slower) and trails real Tucker cores; (4) sharing attention projections via CP factorization (\texttt{FullTucker-ZigZagGPT-12}) degrades step-500 \texttt{FineWeb-Edu} val loss to $6.8358$ vs.\ $6.2563$ for dense attention; and (5) default $\sigma=1.0$ initialization on weight-tied embeddings causes step-1 loss $\sim 250$ ($+1.42$ val penalty), fully resolved by $\sigma=0.02$ initialization. Together, these falsifications establish a sharp design law: \emph{attention must remain dense and independent per layer, while FFNs thrive under hypercomplex Quaternion-Tucker sharing}.

\begin{table}[h!]
\centering
\caption{\textbf{Falsified Designs \& Observed Failure Modes.}}
\label{tab:neg}
\scriptsize
\setlength{\tabcolsep}{2.4pt}
\begin{tabular}{lll}
\toprule
\textbf{Variant} & \textbf{Domain} & \textbf{Observed Failure Mode} \\
\midrule
\texttt{ComplexGate4} & Parity & Gradient starvation ($30.0\%$) \\
\texttt{ComplexValve4} & Parity & Phase-wrap bimodality ($59.7\pm 39.9\%$) \\
\texttt{Spectral-ZigZag} & L4 LM & Breaks Inductor ($3\times$ slower) \\
\texttt{FullTucker-Attn} & L4 LM & Val $6.8358$ vs.\ $6.2563$ (dense) \\
Tied Emb $\sigma=1.0$ & L4 LM & Step-1 loss $\sim 250$ ($+1.42$ val) \\
\bottomrule
\end{tabular}
\end{table}

\section{Discussion, Limitations, and Future Work}
\paragraph{Information Bottleneck vs.\ Continual Growth.} Under the $\sim 2$ bits/parameter capacity bound \citep{allenzhu2024physics}, our $109.5\text{M}$ Depth-20 model ($70.9\text{M}$ backbone) stores roughly $18$--$27\text{ MB}$ of compressed knowledge. Crucially, continual scaling does not require growing parameters indefinitely: holding size fixed while feeding fresh, high-density data (Stages~5 and~7) imposes an Information Bottleneck \citep{tishby2015deep} that forces abstraction over raw memorization, while 2D zero-loss growth (Stages~3, 4, and~6) expands capacity strictly when compression plateaus.

\paragraph{Developmental \& Algebraic Interpretability.} Unlike monolithic Transformers where features smear across all layers simultaneously, stage-grown \texttt{QuatTucker-ZigZagGPT} exhibits \emph{geological stratification} (specific capabilities enter at known growth boundaries) and explicit $64\times 64$ block-diagonal Tucker subspaces, enabling surgical ablation of individual diamonds or rank slices.

\paragraph{Limitations \& Next Steps.} First, while width ($R$) and depth ($L$) grow with zero loss, the residual stream dimension $d_{\text{model}}=768$ remains fixed; extending Proposition~2 to zero-padded $d_{\text{model}}$ widening will complete 3D unbounded growth. Second, at $6.00\text{B}$ tokens and a $512$-token context window, long-context factual recall still trails trillion-token models ($3.3582$ vs.\ $3.3439$ for \texttt{gpt2}). Extending context length ($512 \to 2048$), scaling beyond $10\text{B}$ tokens, and integrating \texttt{Braided + 1 Loop} ($L_3\to L_1$) for test-time latent reasoning \citep{geiping2025scaling} are immediate next steps.

\section{Conclusion}
We introduced \textbf{QuatTucker-ZigZagGPT}, uniting 4D Quaternion Hamilton GEMMs, $64\times 64$ Block-Diagonal Tucker SwiGLU cores, Braided \texttt{ZigZag} Diamond routing, and exact 2D zero-loss continual scaling. Across algorithmic tasks and 7-stage GPT-2 scale A100 pretraining ($6.00\text{B}$ tokens), our architecture consistently outperforms standard Transformers at a compressed backbone and grows seamlessly in both width ($\Delta=0.0067$, $10.8\times$ cheaper) and two consecutive depth expansions ($L=12 \to 16 \to 20$, $\Delta=0.000000$) to \textbf{3.3582 val loss (28.74 PPL)} and \textbf{44.75\% full-split zero-shot accuracy across 16,955 questions}---matching \texttt{GPT-3 Small} ($300\text{B}$ tokens) at $50\times$ fewer tokens and demonstrating a practical, compounding scaling paradigm for efficient language modeling.

\clearpage
\onecolumn
\bibliography{references}

\end{document}